\documentclass[12pt]{article}
\usepackage{amsmath}
\usepackage{times}
\usepackage{graphicx}
\usepackage{xcolor}
\usepackage{multirow}
\usepackage{framed}

\usepackage{rotating}
\usepackage{bbm}
\usepackage{latexsym}
\usepackage{hyperref}
\newcommand{\appropto}{\mathrel{\vcenter{
  \offinterlineskip\halign{\hfil$##$\cr
    \propto\cr\noalign{\kern2pt}\sim\cr\noalign{\kern-2pt}}}}}

\begin{document}
\thispagestyle{empty}
\def\YH#1 {{\bf x}_{\it {#1}}}
\def\YI#1 {{\bf u}_{\it {#1}}}
\def\YO#1 {{\bf o}_{\it {#1}}}
\def\YT#1 {{{\bf u}^{\sf teach}_{\it {#1}}}}

\def \ti {\it t}
\def \tanh {\sf tanh}
\def \sin {\sf sin}
\def \cos {\sf cos}
\def \artanh {\sf artanh}

\def\WWHH {{\bf W}}
\def\WWIH {{\bf w}^{in}}
\def\WWHO {{\bf w}^{out}}

\def\itentitym {{\bf 1}}

\def\gfixp{{{\bf x}_{\infty}}}

\begin{framed} 
\definecolor{blu-gree}{rgb}{0.,.5,.25}
\centerline{\color{blu-gree} \scriptsize  \sf This is a preprint from arxiv.org:1404.6334}
\centerline{\color{blu-gree} \scriptsize  \sf Final version has been published in Neural Computation, 27(5), pp 1102 -- 1119, 2015 (May).}
\centerline{\color{magenta} \scriptsize  \sf Please download the final version from the MIT Press Webpage}
\centerline{\url{http://www.mitpressjournals.org/loi/neco}}
\end{framed}

\vspace{10mm}

{\LARGE \sf Input anticipating critical reservoirs show power law forgetting of unexpected input events}

\ \\
{\bf \large Norbert Michael Mayer$^{\displaystyle 1}$}\\
{$^{\displaystyle 1}$Department of Electrical Engineering and\\
Advanced Institute of Manufacturing with High-tech Innovations (AIM-HI),\\
National Chung Cheng University, Min-Hsiung, Chia-Yi, Taiwan\\
mikemayer@ccu.edu.tw}\\
%

{\bf Keywords:} Reservoir computing, scale invariance, power law

\markboth{}{NC instructions}
\ \vspace{-0mm}\\
%
\begin{center} {\bf Abstract} \end{center}
Usually, reservoir computing shows an exponential memory decay. This paper investigates under which circumstances echo state networks can show a power law forgetting. That means traces of earlier events can be found in the reservoir for very long time spans. Such a setting requires critical connectivity exactly at the limit of what is permissible according the echo state condition. However, for general matrices the limit cannot be determined exactly from theory. In addition, the behavior of the network is strongly influenced by the input flow. Results are presented that use certain types of restricted recurrent connectivity and anticipation learning with regard to the input, where indeed power law forgetting can be achieved.



\section{Introduction}
Stability of a  dynamical system is usually described in terms of the 
eigenvalues of the Jacobian, which clearly can identify a system as 
asymptotically stable or unstable. In the simplest scalar case one might 
consider an iterative system $x_{t+1} = f(x_t)$,
with a fix-point at $x_\infty=f(x_\infty)$.
As a standard approach (see e.g. \cite{invitationdyn1993}) one can analyze the asymptotic behavior by investigating a polynomial series of $f$ around $x_\infty$
\begin{equation}
f(x) \approx x_\infty + \alpha (x-x_\infty) + \beta (x-x_\infty)^2 + \gamma (x-x_\infty)^3 \dots \;.
\label{basics_equ}
\end{equation}

Near the fix-point, one only has to consider the absolute value of $\alpha$ in order 
to predict the future of the dynamics as
\begin{equation}
x_t-x_\infty \propto \alpha^t. 
\end{equation}

Thus, $x_t$ converges to $x_\infty$ if $|\alpha|<1$, and it diverges from $x_\infty$ if $|\alpha|>1$.
However, for the narrow class of 
systems of $|\alpha| =1$ the linear term does not contribute to the convergence or divergence of $x_t$
\footnote{\normalsize This situation is also known as one of the cases where linear stability analysis fails (cf. \cite{invitationdyn1993}, p. 93 ff.) }. 
In this case the system can
be considered critical in the sense that its
dynamical properties can change if relatively small variations of $\alpha$ occur.
Also,
in the critical 
regime the dynamical system is controlled by the higher order terms ($\beta$, $\gamma$ ...) of eq. \ref{basics_equ}.
Among other consequences also the temporal convergence
 can then be approximated by a power law:

\begin{equation}
x_t-x_\infty \approx c \; t^b,
\end{equation}
where $c$ and $b$ are constant values. \\

More general, 
power law properties in the dynamic variables and statistics can appear
in critical states \cite{1993phasetrans}. Vice versa, if power law statistics appear in 
experimental measurements, often a dynamical system in a critical regime can be 
assumed \cite{prl1987}.
For example,
in brain slices of rats, electrode measurements revealed 
such kind of statistics in spontaneous activity cascade lengths (\textquoteleft avalanches\textquoteright) \cite{slices}. 
If networks in the brain are dynamical systems in a critical regime
this is an interesting feature which 
raises the question of the 
biological purpose of such a phenomenon. \\

\begin{figure}[t]
\begin{center}
\includegraphics[  width=0.38\paperwidth]{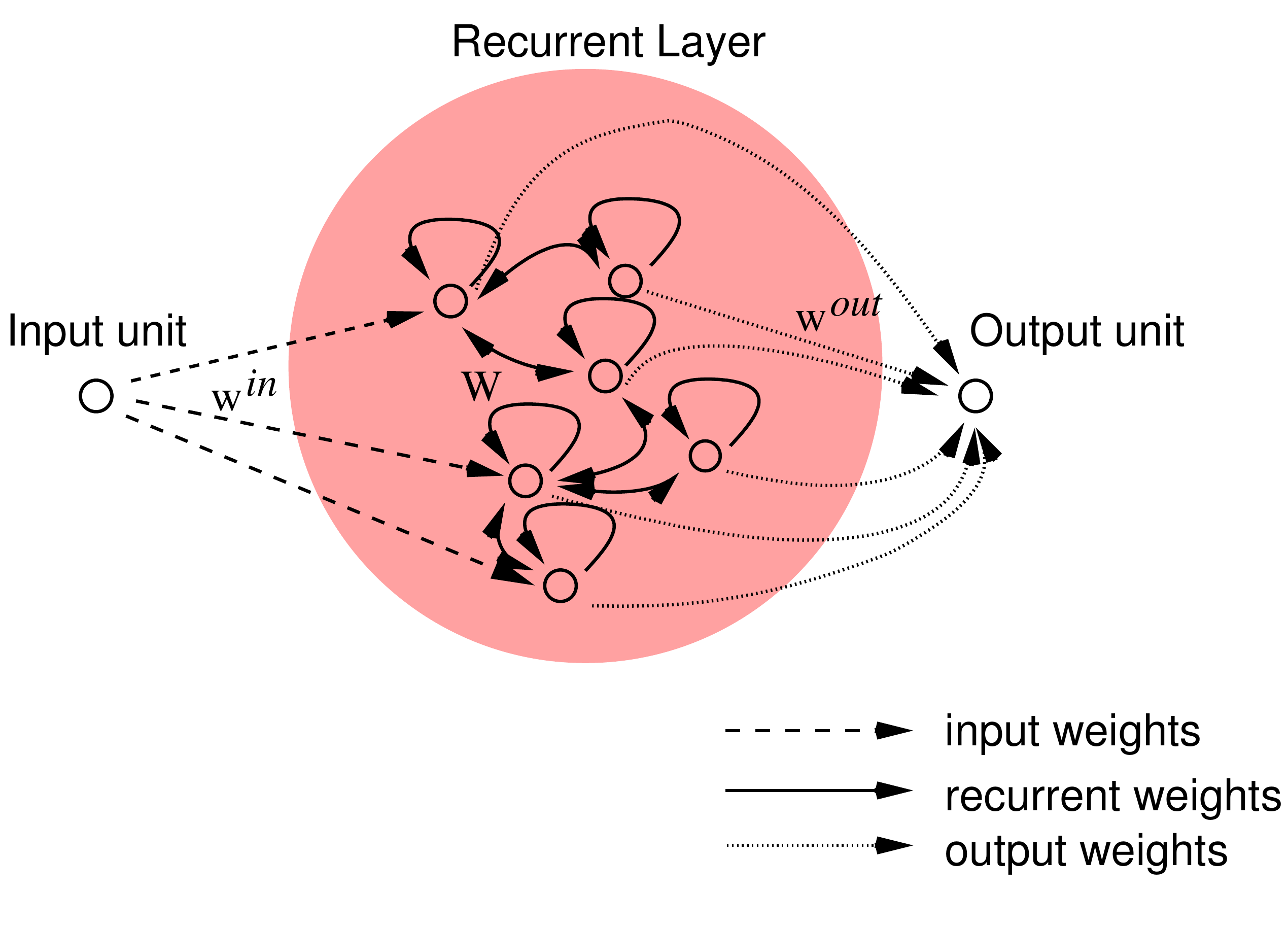} 
\end{center}
\caption{\label{scheme} ESN scheme: The initial ESN approach chooses input and recurrent connectivity randomly, although it has to obey the echo state condition. 
Learning is only applied to the output layer.
What is the {\em best} connectivity with regard to the other layers and some certain input statistics is still subject to an ongoing debate to which this work would like to contribute. 
 }
\end{figure}

In this paper a reservoir computing approach (for an overview over some recent developments see
\cite{special_issue}) is presented where a critical dynamics is arranged around the expected input, that is 
if the network is trained with the expected input, it forgets any unexpected input in a power law fashion.
Work of  
\cite{danko} can be seen as a hint 
that at least with respect to untrained stimuli, some cortical areas 
can be interpreted as a kind of reservoir computing.  
In the scope of this work, I will introduce a neural networks approach 
that is based on Echo-State-Networks (ESNs). 
The ESN is a reservoir approach that uses rate-based neurons with real valued transfer functions, initially sigmoid functions.
For my very basic considerations it turned out to be sufficient and can be easily tuned into a
critical regime in the sense of this paper. 
The most important foundations of this paper's approach are those approaches that relate to critical ESNs \cite{criticalESNs2006,theobi2012,boeobs2014,2010buesing, 2005natschlaeger}.\\

The learning approach is specifically designed in a way that the network learns to anticipate the input, i.e. to predict it. 
This anticipation can also be labeled as balancing the network, a term which has 
been coined earlier \cite{balancednn} in the context of chaotic dynamics of 
integrate and fire neurons. 
The purpose of this mechanism is to prevent predictable information from entering 
the network, and producing activity in the recurrent layer. The intention
is rather to let only the un-predicted input, i.e. special events with some level of unusualness, pass into the recurrent layer 
and stay there for further processing, while standard ESN approaches can only keep information about these events in the order of the half-time period of the forgetting factor.  
\\

In the following section I introduce the Echo-State paradigm and how 
it can be tuned 
into the critical state. I then discuss the role and implementation of
 the anticipation/balancing 
part of the algorithm. A result section follows. I conclude the work 
with a discussion concerning the potential information theoretic benefit of my model and biological implications.

\section{Model}
The model is based on J\"ager's ESN approach \cite{jaeger1,jaeger2}. It consists of an input, recurrent layer and possibly an output layer (though not explicitly introduced in the scope of this paper). ESNs are composed of rate-based neurons with real valued transfer functions; the update rule is:

\begin{eqnarray}
\YH { lin, \ti} &=& { \WWHH \YH { \ti-1 } } + \WWIH \YI { \ti } \label{xlin}\\
\YH { \ti} &=& \theta \left(   \YH { lin, \ti}     \right)  \\
\YO { \ti} &=& \WWHO \YH {\ti }
\label{hidden_dyn} 
\end{eqnarray}
where the vectors $\YI { \ti } $,$ \YH { \ti } $, and $ \YO { \ti } $ are the input, the neurons of the 
hidden layer, and the neurons of the output layer, respectively, and $\WWIH$, $\WWHH$, and $\WWHO$ are the matrices of the respective 
synaptic weight factors. 

As a convention the transfer function $\theta(.)$ is continuous, differentiable and monotonically increasing with the limit
$1 \geq \theta'(.) \geq 0$, which is compatible with the requirement that $\theta(.)$ fulfills the Lipschitz continuity with $L=1$.

 J\"ager's approach uses random matrices for $\WWHH$ and $\WWIH$, learning is restricted to the output layer $\WWHO$ (see fig. \ref{scheme}). 
The learning (i.e. training $ \YO { \ti } $) can be performed by linear regression.
 Since the learning process with regard to some output itself is not of interest in the scope of this paper, this part of the approach is not outlined here.

\subsection{Echo-state condition}
A necessary condition for the performance of an ESN network is that the echo state condition is fulfilled.
Consider a time-discrete recursive function ${\bf x}_{\ti +1} = F({\bf x}_{\ti  }, 
{\bf u}_{\ti } )$, where the  
${\bf x}_{\ti }$s are interpreted as internal states and the ${\bf u}_{\ti }$s form some external input sequence, i.e. the stimulus.
The definition of the echo-state condition is the following.
Assume an infinite stimulus sequence: $\bar{\bf u}^{\infty} = {\bf u}_0, {\bf u}_1, \dots$ 
and two random initial internal states of the system ${\bf x}_0$ and ${\bf y}_0$. 
To both initial states ${\bf x}_0$ and ${\bf y}_0$, the sequences 
$\bar{\bf x}^{\infty} = {\bf x}_0, {\bf x}_1, \dots$ and $ \bar{\bf y}^{\infty} = {\bf y}_0, {\bf y}_1, \dots$ can be assigned; 
\begin{eqnarray}
{\bf x}_{\ti+1} &=& F({\bf x}_{\ti },{\bf u}_{\ti } ) \label{sesn1} \nonumber \\
{\bf y}_{\ti+1} &=& F({\bf y}_{\ti },{\bf u}_{\ti } ) \label{sesn2}.
\end{eqnarray}
Then the system $F(.)$
is called {\em universally state contracting} if {\bf independent} from the 
set ${\bf u}_{\ti }$ and for {\bf any} 
(${\bf x}_{0}$,${\bf y}_{0}$) and all real values $\epsilon > 0$,
there exists  an iteration $\tau$ for which 
\begin{equation}
d({\bf x}_{\ti }, {\bf y}_{\ti }) \leq \epsilon \label{metric}
\end{equation}
for all $ \ti  \geq \tau$. \\
J\"ager showed  (\cite{jaeger1} pp. 43) that the echo state condition is fulfilled if and only if the network is universally state contracting.
The ESN is designed to be universally state contracting and thus to fulfill  the echo-state condition. Heuristics show that the
performance of ESNs becomes better near the critical point where the echo state condition is just narrowly fulfilled. The rest 
of this paper is dedicated to determining under which circumstances the time series of eq. \ref{metric} can converge as
\begin{equation}
d({\bf x}_{\ti }, {\bf y}_{\ti }) \appropto {\ti}^{b},
\end{equation} 
rather than the usual
\begin{equation}
d({\bf x}_{\ti }, {\bf y}_{\ti }) \appropto {a}^{\ti},
\end{equation} 
which connects reservoirs to the considerations from the introduction and eq. \ref{basics_equ}, i.e. to critical connectivity.

Critical connectivity can be achieved by just narrowly fulfilling J\"ager's conditions on the recurrent weight matrix of a network that has echo states:\\
A necessary condition is: 
\begin{itemize}
\item {\bf \sf C1} A network has echo states only if the absolute value of the biggest eigenvalue of $\WWHH$ is below $1$.
\end{itemize}
A sufficient condition is: 
\begin{itemize} 
\item {\bf \sf C2} 
A network has echo states if the  biggest  singular value of $\WWHH$ is smaller
than one\footnote{\normalsize A closer sufficient condition has been found in \cite{tighter,tighter2}. It is slightly better than the one outlined here, but still leaves a gap to the necessary condition for general matrices.}. 
\end{itemize}

\subsection{Normal recurrent connectivity matrices $\WWHH$}
A general ESN becomes critical -- for some input sequences -- somewhere in the range between {\sf C1} and {\sf C2}. 
In the case of { \bf normal} matrices for the recurrent connectivity $\WWHH$, i.e. matrices that fulfill:
\begin{equation}
\WWHH^T \WWHH = \WWHH \WWHH ^T, \label{normal_matrix}
\end{equation}
both conditions are either true or false at the same time. The proof can be found for example 
in \cite{1991spctsglvalues}.

Several prominent types of matrices are normal: symmetric, orthogonal(${\cal O} (n)$), permutation and skew symmetric.
 
\begin{figure}[t]
\begin{center}
\includegraphics[  width=0.28\paperwidth]{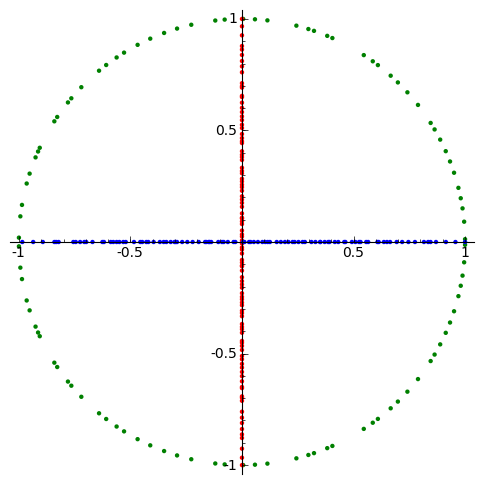} 
\end{center}
\caption{\label{eigenvalue} 
Plot of eigenvalues of samples of different types of normal matrices are depicted according to their positions on the complex plane (x represents the real part, y the imaginary part), 
the eigenvalues of a symmetric matrix are blue, eigenvalues of a skew-symmetric matrix are red and eigenvalues of an orthogonal matrix are green. }
\end{figure}

These different types have advantages and disadvantages with respect to learning algorithms:
First of all permutation matrices can be discarded:
Although untrained reservoirs of permutation matrices tend to show a good performance compared to other types of connectivity patterns,
they have a significant disadvantage with regard to optimization tasks. For a recurrent layer with $n$ neurons there are $n!$ different 
permutations possible, which forms a discrete set of disjunct points rather than a continuous manifold within the parameter space. So 
they seem hardly usable here since fine tuning cannot be performed by continuous parameter optimization.
Symmetric matrices are less useful since such reservoirs show only trivial dynamics in real valued reservoirs, i.e. the eigenvectors are mapped to themselves.
Orthogonal matrices are a good choice. They form a group with respect to matrix multiplication, their absolute eigenvalues are all $1$
(see fig. \ref{eigenvalue}); the complete set of all orthogonal matrices forms a manifold that can be parameterized. Special virtues of 
orthogonal matrices have already been investigated \cite{PhysRevLett.92.148102} in a different context.
Skew-symmetric also might be useful. This may be biologically plausible as a result of STDP \cite{markram2012spike}.
However, considering fig. \ref{eigenvalue} one can see that the eigenvalues are all pure imaginary 
and more or less equally distributed along the imaginary axis. That is why -- similar to symmetric matrices -- most components of the signal encoded in the hidden neurons decay exponentially and only the information along the eigenvectors belonging to the two largest eigenvalues ($\pm i$) survive for a power law time scale. Thus, since usually only two eigenvectors have the highest absolute eigenvalue, the information carried in the critical state might be poor.
In the scope of this work, orthogonal matrices are being investigated. Previously also skew-symmetric matrices have been checked.
Technically, one can obtain a skew-symmetric or orthogonal matrix by adding additional 
steps into the learning process that -- in every iteration -- enforce the respective constraints. 

In the present case one has to enforce either one of the two 
following constraints 
\begin{itemize}
\item {\bf \sf C3} skew-symmetric $\WWHH$ or 
\item {\bf \sf C4} orthogonal $\WWHH$.
\end{itemize}
In the case of skew-symmetric matrices, one also has to set
\begin{itemize}
\item {\bf \sf C5} largest absolute eigenvalue/singular value is set to $1$. 
\end{itemize}
Orthogonal matrices already have all their absolute eigenvalues set to $1$.
Thus, if the network has to be tuned exactly into the critical state
one way to do that is to either enforce {\sf C3} and {\sf C5} or {\sf C4} in every
iteration. For practical purposes it suffices in many cases to use
general matrices with a largest absolute eigenvalue set to one (i.e.
only enforce condition {\sf C5}).

\subsection{Transfer function and anticipation \label{anticip_sect}}

In order to identify critical behavior, one has to come back to eq. \ref{metric}, which basically defines convergence between two different state sequences if the input is identical and the network connectivity is also identical. The most direct way to enforce the convergence is to design the network as contractive, i.e. to ensure in each iteration $t$:
\begin{equation}
d({\bf x}_{t+1}, {\bf y}_{t+1}) = {l}(t) \cdot d({\bf x}_{t}, {\bf y}_{t}) \label{ldef1},
\end{equation}
where $l(t)$ indicates whether the distance between ${\bf x}_{t}$ and ${\bf y}_{t}$ converges to zero.
If $l(t)$ becomes and stays equal or larger than $1$ from any $t$, the two sequences do not converge anymore and the network does not have the echo state property. On the other hand if for all $t$ and an upper limit $L$ one has the inequality
\begin{equation}
l(t) \leq L < 1,
\end{equation} 
then network is an ESN.

Eq. \ref{ldef1} defines $l(t)$ as 
\begin{equation}
l(t) = \frac{d({\bf y}_{t+1}, {\bf x}_{t+1})} {d({\bf y}_{t}, {\bf x}_{t})} =
\frac{|| \theta(\WWHH {\bf y}_t + I) - \theta(\WWHH {\bf x}_t + I) ||_2} {|| {\bf y}_{t} - {\bf x}_{t} ||_2 },
\end{equation}
where $I=\WWIH \YI { \ti } $ is the influx from the input to the network, and $\left||.\right||_2$ is the Euclidean norm.
For tiny differences between ${\bf x}_t$ and ${\bf y}_t$ one can approximate $l(t)$ as
\begin{equation}
l(t) \approx   
\frac{||\dot{\Theta} \WWHH ({\bf y}_t- {\bf x}_t)||_2} {||{\bf y}_t- {\bf x}_t||_2} \leq || {\bf J}||_2,
\end{equation}
where $\dot{\Theta}$ is a diagonal matrix with entries
$\dot{\theta}({\bf W} x + I)$, and the right hand expression is the spectral norm of the Jacobian of the neural network dynamics, i.e.
\begin{equation}
J_{ij} = \dot{\theta} (x_{lin,i}) W_{ij}，
\end{equation}
where $x_{lin,i}$ is the ith component of the vector $\YH { lin, \ti} $ from eq. \ref{xlin}.
Note that by definition of the spectral norm, $l(t)$ approximates 
$|| {\bf J}||_2$ if the difference between ${\bf y}_t $ and ${\bf x}_t$ is co-linear to
the largest eigenvector of $\bf J$.
 
With respect to criticality the choice of the transfer function is important. 
On one hand note that the ESN requires convergence for {\bf any} input sequence which explains why $1 \geq \dot{\theta}$. 
On the other hand, if one can achieve 
\begin{equation}
1 = \dot{\theta}(x_{lin,i}),
\label{crit_sigmoid}
\end{equation}
one gets simply $\bf J = \WWHH$, and thus $|| \bf J ||_2=1$. This case is analogous to the situation of eq. \ref{basics_equ} and the case $|\alpha|=1$ there; the linear terms do not affect the dynamics anymore. 
Instead, the dynamics are rather controlled by higher order terms of Taylor series around the critical point. The shape of the higher terms depends on the general shape of the transfer function; see 
\cite{proof2014} on an analysis of critical state, in particular under which circumstances the echo state condition can be preserved.

Then one fundamental problem of creating critical behavior is to design the network in a 
way that the $x_{lin,i}$s are always tuned to values where eq. \ref{crit_sigmoid} is fulfilled.
The way to this is to let the network anticipate, i.e. predict the next input. Ideas that lead to 
the prediction of input have been outlined in earlier works \cite{BIOADIT2004,jjsteil2012,2012sussiloabbott}. \\

Usually $\theta=\tanh(.)$ is used,  and it is used in the initial approach. 
Note that the critical point of eq. \ref{crit_sigmoid} is reached at 
\begin{equation}
\YH { lin, \ti} = [ 0, 0, 0 \dots ],
\end{equation} 
i.e. there is only one possible critical state. This is un
favorable as one can see from the following considerations.

The optimization of the hidden layer is going to redirect the linear response resulting from the expected input exactly to the vector above. Since in the case of the 
hyperbolic tangent there is only one possible value, i.e. all linear responses are $0$.

On the one hand, the optimization process will result in the same single response of the entire network for all expected inputs.
On the other hand, the point of the recurrent network is to resolve ambiguities by accounting for the input history. This is not possible if the total network is indifferent to what the previous input was if this input was expected. 

So there is no information transfer from the previous
state to the next state.
For the current purpose, it makes sense to use the following transfer function:

\begin{equation}
\theta(x) = 0.5 x - 0.25 \, {\sin}(2 x) 
\label{sigmoid}
\end{equation}
It has to be noted that the maximal derivative $\theta'(x)$ is $1$ at 
$x_{lin, i} = \pi(n+1/2)$, where $n$ is an integer number. 
Usually for small input values, only the lowest states 
$x_{lin, i} = \pm \pi/2$ are used by the network, resulting in a 
 setting resulting in a setting where the critical point of eq. \ref{crit_sigmoid} is reached as
\begin{equation}
\YH { lin, \ti} = [ \pm \frac{\pi}{2}, \pm \frac{\pi}{2}, \pm \frac{\pi}{2} \dots ].
\end{equation} 
This results in $2^N$ different possible states for the total network, where $N$ is the number of the hidden neurons. Thus, information transfer within the critical state is possible.

\subsection{Learning anticipation in the hidden layer}
Several learning rules for the hidden layer have been proposed. 
In many cases an information theoretic measure has been applied \cite{boeobs2014,jjsteil2006}. 
In the present approach it is intended to follow the idea to anticipate the input, which also has been tried in different ways previously \cite{criticalESNs2006}. \\

I propose here the minimization of the cost-function
\begin{equation}
E(\WWHH, \WWIH ) = \sum_i \, <cos(\YH { lin, {\ti,i}} )^2>_t,
\label{costf}
\end{equation}
where $\YH { lin, {\ti,i}} $ is the $i^{th}$ component of $\YH { lin, \ti } $ (see eq. \ref{xlin}). 
The cost function becomes minimal if $ \YH { lin, {\ti,i}} = \pi(n+1/2)$, which fits to the critical points of the transfer function eq. \ref{sigmoid}.
Note that $\YH { lin, \ti } $ also contains input to the system at the time $\ti$. Thus, the optimization includes an \textquoteleft anticipation\textquoteright  of the input, i.e. some input is expected and already counted in order to come close to $\pm\pi/2$ for each neuron. The optimization is done on both $\WWHH$ and $\WWIH$ by gradient descent
\begin{eqnarray} 
\Delta \WWHH &=& - \nabla_{\WWHH} \; E  \nonumber \\
\Delta \WWIH &=& - \nabla_{\WWIH} \; E,
\end{eqnarray}
where $E$ is the cost-function of eq. \ref{costf}, i.e. one gets
\begin{eqnarray}
\Delta \WWHH _{ij} &=& 
-2 \; {\cos} (\YH { lin, \ti,i} ) \; {\sin} (\YH { lin, \ti,i} ) \, \YH{\ti-1,j} 
\nonumber \\
\Delta \WWIH _{ij} &=& 
-2 \; {\cos} (\YH { lin, \ti,i} ) \; {\sin} (\YH { lin, \ti,i} ) \, \YI{\ti,j}.
\end{eqnarray}
The update is done by 
\begin{eqnarray}
\tilde{\WWHH} &=& \WWHH_{\sf old} + \epsilon \Delta \WWHH \nonumber \\
\WWIH_{\sf new}         &=& \WWIH_{\sf old} + \epsilon \Delta \WWIH,
\end{eqnarray}
where $\WWHH_{\sf old}$ and $\WWIH_{\sf old}$ are recurrent and input connectivity matrices of the previous iteration, $\epsilon$ represents the learning rate, and 
$\WWIH_{\sf new}$ is the updated input connectivity matrix. 

As a result of the learning process the updated $\tilde{\WWHH}$ is not necessarily critical, respectively normal in the sense of eq. \ref{normal_matrix} anymore. Thus, it remains to find a normal matrix that is as near as possible to
the result of the learning step.
Thus,  in each iteration after adaptation the conditions {\sf C4} and {\sf C5} can be enforced by setting
\begin{eqnarray}
[{\bf U},{\bf S},{\bf V}^T] &=& {\sf SVD} ( \tilde{\WWHH}) \nonumber \\
\WWHH_{\sf new} &=& {\bf U} \cdot {\bf V}^T, \label{constraint_enf}
\end{eqnarray}  

where ${\sf SVD}$ represents the singular value decomposition routine of standard mathematical toolboxes, usually the return value is a list composed of three matrices, i.e. 
$[{\bf U},{\bf S},{\bf V}^T]$ where ${\bf U}$ and ${\bf V}$ are two orthogonal matrices and 
${\bf S}$ is a positive semi-definite diagonal matrix. Since the constraint is enforced in every iteration, already $\WWHH_{\sf old}$ is an orthogonal matrix. After the learning step, if we assume 
$\epsilon$ to be small, one can expect ${\bf S}$ to be near an identity matrix. 
Setting $\WWHH_{\sf new}$ according to eq. \ref{constraint_enf} makes it an orthogonal matrix since the right hand side of eq. \ref{constraint_enf}  is a product of 2 orthogonal matrices. So the recurrent connectivity can forced back on the manifold of orthogonal matrices. 
In this way condition {\sf C4} is enforced.
Since all absolute eigenvalues of orthogonal matrices 
are equal to one, the condition {\sf C5} is also enforced by eq. \ref{constraint_enf}.
For symmetric matrices as a different type of normal matrices other ways of enforcing constraints can be applied

For example, as outlined in \cite{2014ICANN}, C3 and C4 can be enforced by
\begin{equation}
{\bf W}_{new} = \frac{\tilde{\bf W} - \tilde{\bf W}^T} {\max {\rm abs}({\rm
eigv}(\tilde{\bf W} - \tilde{\bf W}^T))}
\end{equation}
where $\rm eigval$ represents a function that computes numerically a
list of eigenvalues of a matrix.
In the case of general matrices C5 can approximately enforced by

\begin{equation}
{\bf W}_{new} = \frac{\tilde{\bf W}} {\max {\rm abs}({\rm eigv}(\tilde{\bf W}))}.
\end{equation}

\begin{figure}[t]
\begin{center}
\includegraphics[  width=0.28\paperwidth]{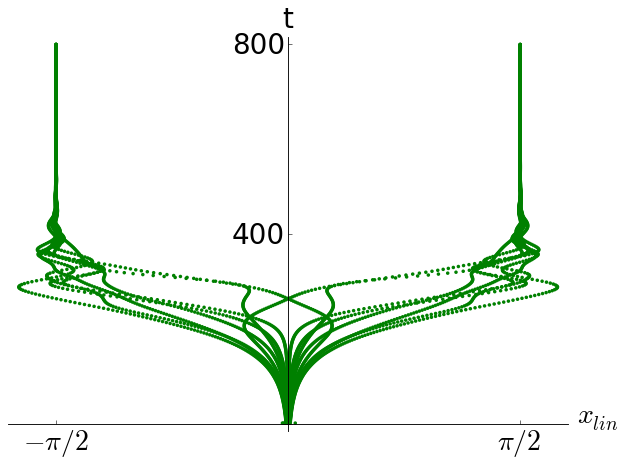} 
\includegraphics[  width=0.28\paperwidth]{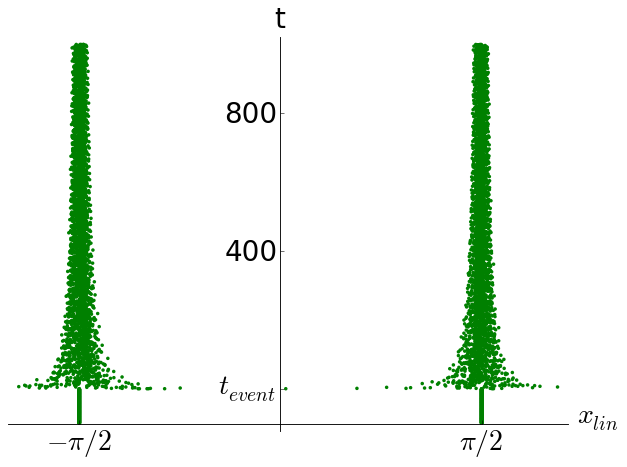} 
\end{center}
\caption{\label{linear_resp} 
Linear response of the 2 state test setting.
Left: Linear response during the learning process.
Right: Linear response of network after an un-expected input event. One can see slow convergence process to the target values $\pm \pi/2$. 
}
\end{figure}

\section{Simulation results}

\begin{figure}[t]
\begin{center}
\includegraphics[  width=0.28\paperwidth]{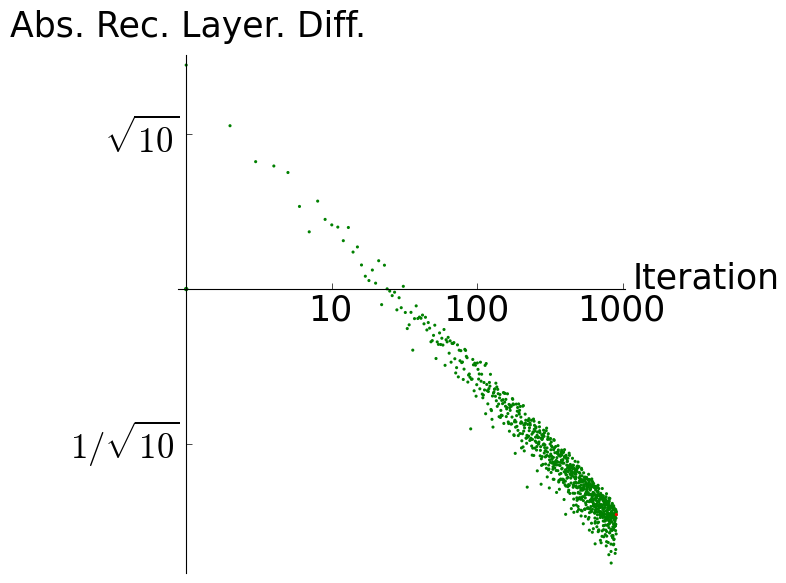} 
\includegraphics[  width=0.28\paperwidth]{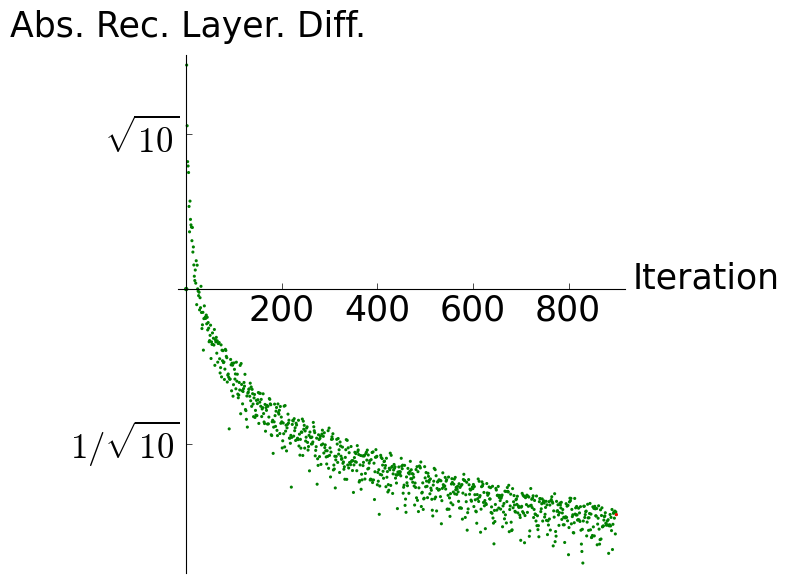} 
\end{center}
\caption{\label{measure} 
Left: Double log plot of the reduced model. The x-axis depicts time (iterations), and the y-axis depicts the difference measure between the 
undisturbed network with a network that received one unexpected input at one iteration. 
One can see that the decay is proportional to $t^a$, which results in a straight line of points in the double log plot. Iterations are counted from the moment at which one network receives one unexpected input. Right: Log-lin plot of the same data.}
\end{figure}

\subsection{Reduced model}
The first experiment was done with a model of 8 neurons and a normalized orthogonal connectivity matrix in the recurrent layer. The input was one neuron alternating between $1$ and $-1$. In this case the learning algorithm for the hidden layer converges rapidly 
($\approx$ 5000 iterations with a learning rate of 0.01). 
Fig. \ref{linear_resp} depicts on the left side the linear response of each neuron (${\bf x}_{lin,t}$) during the learning process. 
One can see that the learning follows the intended effect, i.e. the linear responses become more and more accurately either $-\pi/2$ or $\pi/2$. Note that with different initializations
other values of the set $\pi/2 + n\times \pi$ can be approximated, where $n$ is an integer number.

As a second step 2 identical copies of the network are created after the learning process. 
Then an unexpected input ($\YI {} =-1$ instead of $\YI {} =1$) is presented to one of these networks at one time step. 
Fig. \ref{linear_resp}, right side, depicts the recording of the linear response of each of the 8 neurons, from briefly before the unexpected input until several hundred iterations afterwards. One can see that, although the neurons converge to either $-\pi/2$ or $\pi/2$, this process is very slow.   

In order to get a quantitative picture, one can also take  a metrical measure, in an analogous fashion to eq. \ref{metric}\footnote{ \normalsize The applied metric was $d=\sum_i {\sf abs} (x_i-y_i)$, where $x$ and $y$ are the respective states of both networks.}.

Results are depicted in fig. \ref{measure}. The results indicate that the reduced model shows the echo state property, however the decay is not exponential but a power law. Therefore, for a very long time period traces of the single unexpected event can be found in the network. Experiments using skew-symmetric matrices 
have been conducted previously and are going to be published in \cite{2014ICANN}

\begin{figure}[t]
\begin{center}
\begin{tabular}[t]{ll}
{\sf Regular expression} & {\sf Breach of rule} \\
\includegraphics[width=0.28\paperwidth]{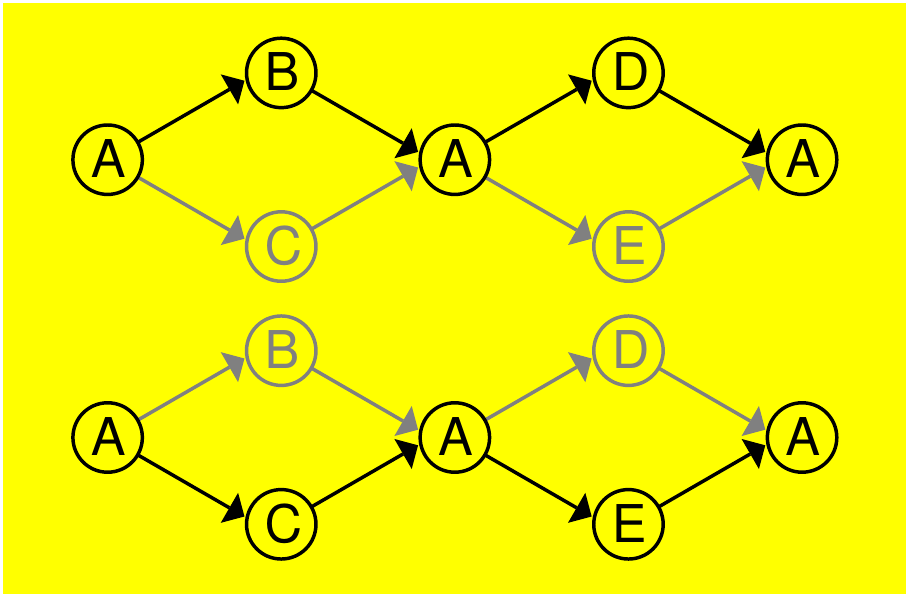} &
\includegraphics[width=0.28\paperwidth]{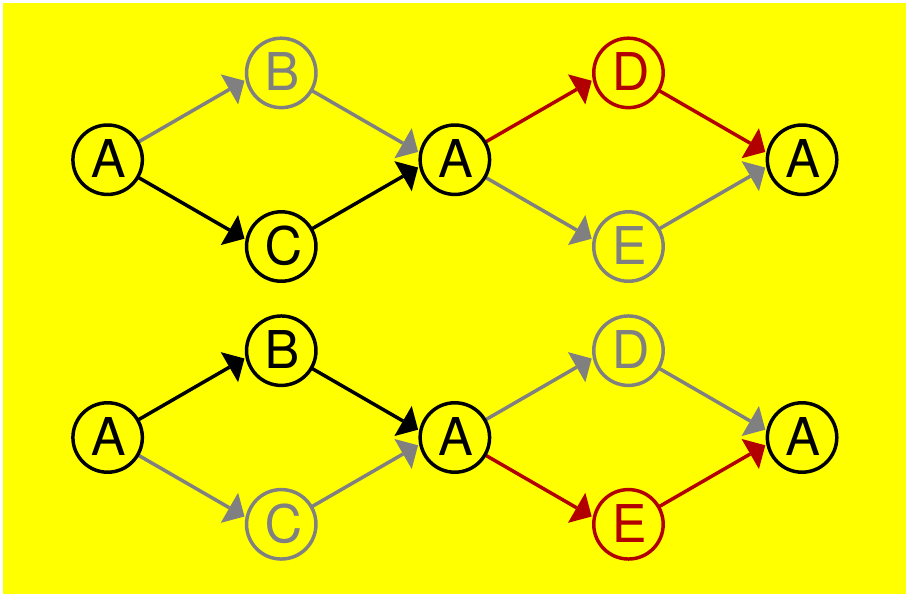} \\
{\sf Sample: ACAEABADA$\dots$} & {\sf ACAEABA{\color{red} E}A$\dots$ }
\end{tabular}
\end{center}
\caption{Grammar of the training set: \label{state_sequence} Left the trained sequence of states. A regular sequence of four steps consists of either the sequence {\sf ABAD} or {\sf ACAE}. 
The occurrence of both sequences is random with 
equal probability. After four steps the sequence is repeated from the beginning again. For testing the networks, 2 identical networks were used where one network at one step received input at one time from a sequence that violates the implicitly trained rule. }
\end{figure}

\begin{figure}[t]
\begin{center}
\includegraphics[  width=0.3\paperwidth]{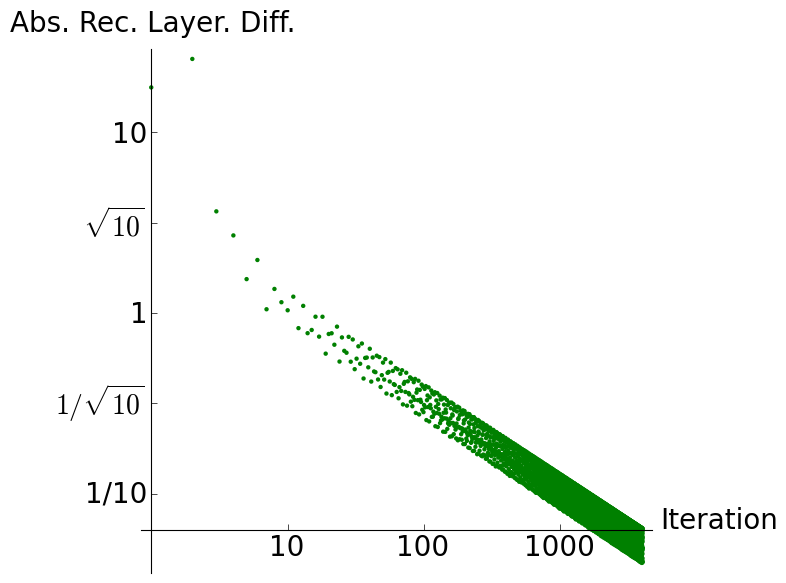} 
\hskip 0.1cm
\includegraphics[  width=0.3\paperwidth]{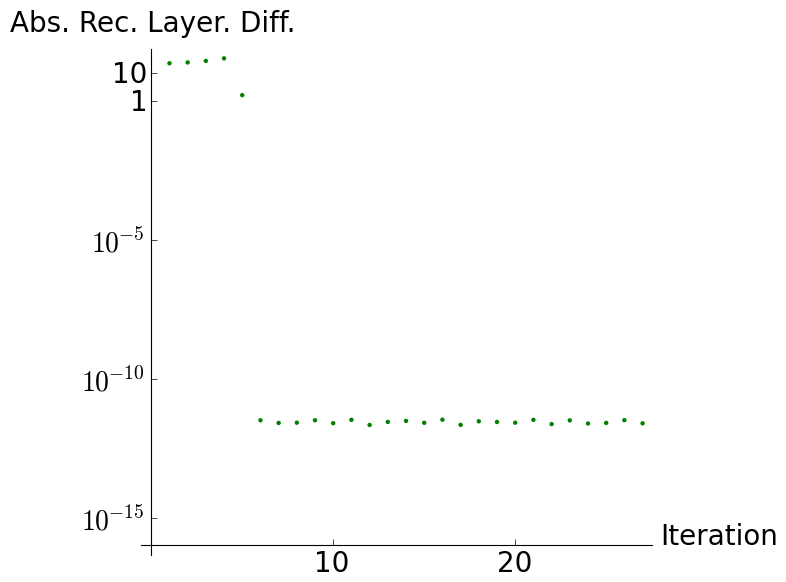}
\end{center}
\caption{\label{singlebreak:nobreak}
The plots show the convergence between two identical networks where one network receives a different input from the other network for a brief time. Left: 
Here one network receives
at one iteration an input that represents a grammatical error in the context of the training set whereas the other network receives the grammatically correct sequence. The plot is depicted in a double logarithmic way, thus revealing power law descent of the metric distance between the two networks. The exponent
of the decay is approximately $-0.5$.
Right:
Semilogarithmic plot of two networks that receive different, however in both cases grammatically correct input,  at 2 time steps.
Except for iteration 1 and iteration 3 both networks receive the same input. Thus, the first network perceives the sequence 
{\sf ABAD}, whereas the second network gets {\sf ACAE}.
}
\end{figure}

\begin{figure}[t]
\begin{center}
\includegraphics[  width=0.38\paperwidth]{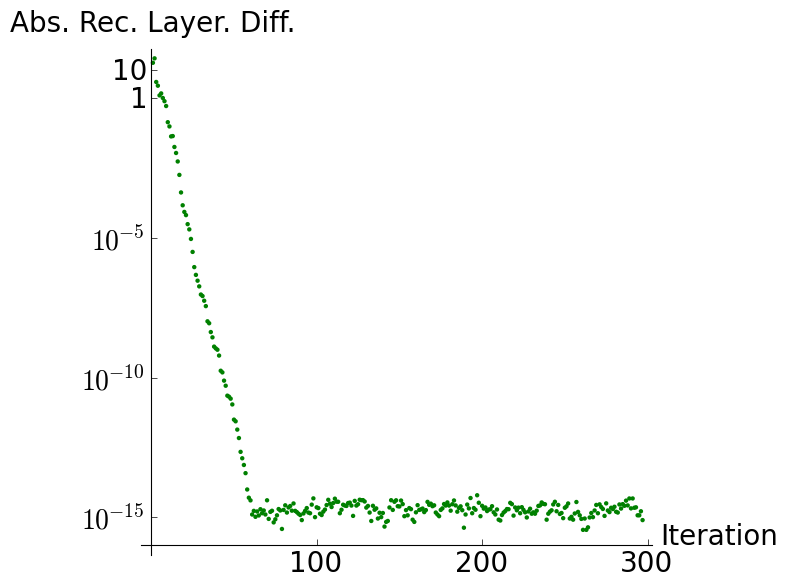}
\end{center}
\caption{\label{permbreak} 
In this plot initially one network is exposed to violations of the grammatical rule, whereas the other network still receives the correct input. After that both networks receive permanent input with the wrong grammar. The plot is log-lin, which demonstrates that the difference between both networks vanishes exponentially in this case.
}
\end{figure}

\subsection{Stochastic sequences with temporal inference}
A similar input model as in \cite{2010c} was chosen to set up a subsequent test. Here an input sequence composed of 5 different state types labeled $\sf A-E$ (see fig. \ref{state_sequence}) was used.
Subsets of 4 subsequent states always belong together.
The subset is either $\sf ABAD$ or $\sf ACAE$, both occur i.i.d. randomly with equal probability. 
The sequence implies a simple grammar, i.e. the occurrence of a $\sf D$ in the state sequence requires the occurrence of a $\sf B$ two iterations before, while the occurrence of a $\sf E$ requires a 
$\sf C$. Thus the network should infer the occurrence of the state $\sf D$ from the previous occurrence of the state $\sf B$ and the occurrence of $\sf E$ from the previous occurrence of $\sf C$.
     
Each of the states are encoded into different input vectors. In the present case a 2 dimensional input vector was implemented to encode each of the five states:

\begin{equation}
\begin{array}{cccccc}
{\sf A} & \rightarrow & ( & 0  &  0  & )\\
{\sf B} & \rightarrow & ( & -1 &  0  & )\\
{\sf C} & \rightarrow & ( &  1 &  0  & )\\
{\sf D} & \rightarrow & ( & 0  & -1  & )\\
{\sf E} & \rightarrow & ( & 0  &  1  & ).
\end{array} \nonumber
\end{equation}

For this experiment general matrices were used and a largest eigenvalue of $1$ was enforced. A network of 15 neurons was capable of virtually reaching the theoretical limit of the cost function during training. 
Instead of training an output function, the present approach again restricts itself to presenting a single unpredicted input and comparing the time series of internal states to an initially identical network where except for the one step over the whole time series, the same input is presented to the network. During the training phase, the network runs 20000 iterations. Initially the recurrent orthogonal matrix is scaled to $S_{max}=0.8$. During the first 7500 iterations $S_{max}$ is exponentially increased to $1.0$. The learning rate of the gradient descent is set to $0.009$. The learning process
usually reaches values of the cost function that are around or smaller than $10^{-20}$. After a transient of 1000 iterations, an exact copy of the first ESN is created. After that three different tests are performed.

In the first test, one network perceives a grammatical error at one iteration (see fig. \ref{state_sequence}, right side), i.e.
both get the same time series as stimuli, except for a single occurrence at which the first networks 
receives an E according to the grammar and the other network receives an $\sf F$, which violates the implicit rule of the training set. Left side of fig. \ref{singlebreak:nobreak} depicts how the difference between the 2 internal state vectors develops over time. 
The first iteration is identical to the iteration where the different inputs occur.
 
As one can see the single violation results in an power law convergence. The metric distance between the two networks is 
\begin{equation}
|| {\bf x}_t -{\bf y}_t || \propto t^{-b},
\end{equation}
where $b$ is approximately $\frac{1}{2}$.
 
In contrast to the first result, if both networks receive 2 different input events that do not represent a violation of the grammar, the convergence of both networks is reached after 5 iterations (see fig. \ref{singlebreak:nobreak}, right side). So in this case the trained network dynamics can return to default after one cycle of training patterns.

Finally, the third experiment tests what happens if both networks receive permanent violations in every iteration after receiving one iteration of differing input. Results are shown in fig. \ref{permbreak}.
As one can see the convergence is exponential in this case. One way to understand this result is to consider the information capacity of the reservoir as limited. As the network is not trained 
to predict the input, the unusual input drives the network further from the critical state.

\section{Discussion}
This paper is intended to contribute to the up-coming discussion about neural networks with critical connectivity \cite{schuster2014}.
Results presented here may contribute on both technical aspects of the study of biological neural networks such as the human cortex.

Technically, a system has been presented in which memories fade slower than exponential. In the present approach, that target could be achieved by applying two components into reservoir computing, i.e.
\begin{itemize}
\item {\bf critical connectivity in the recurrent layer},
\item {\bf an input prediction system}, a system that 
anticipates the next input in a way that it redirects the expected activity 
exactly into the critical point of the system.
\end{itemize}
These two components may also be important features in future approaches of reservoir computing and input driven systems in general.

With regard to the technical aspects of critical connectivity, note no heuristic measure to find the critical point has been applied.This differs from previous approaches that investigated the behavior reservoirs near the edge of chaos \cite{criticalESNs2006,theobi2012,boeobs2014,2010buesing, 2005natschlaeger} . 
 The graphs presented in the results section show a power law convergence over several orders of magnitude, which is only possible if the critical point is hit exactly. On one hand this is presumably not achievable by using the methods applied in earlier approaches. On the other hand, this is necessary to produce the plots that depict power law forgetting over several orders of magnitude. Further research is necessary how the restriction to normal matrices instead of the general ones affects the dynamics in the reservoir in a negative way.

Further, prediction of the input is essential for the results of this paper. Anticipation in reservoirs is possibly an important aspect for the future of ESNs. In addition to the considerations that are outlined in sect. \ref{anticip_sect}, another information theoretic argument supports this notion. 
The idea is here that the total information content of the network is limited. In a non-predictive model the constant information influx requires that knowledge from previous experiences vanishes in an exponential fashion. No matter how the memory is organized, this is inevitable because the capacity of the reservoir is limited.

The predictive/anticipative approach uses a primitive type of memory compression which allows a single unexpected event to stay longer than exponential in the memory. In addition the present work could also be seen as an optimal predicting machine in the sense of Still's definition \cite{PhysRevLett.109.120604}.

Related research can be found in the field of heterogeneous neural models \cite{hoerzer2014, 6795963}, where single input events can also be remembered over long time periods. In the case of LSTM, this can be achieved by adding specific gates in each unit that protect certain information from being forgotten.

Moreover, there could also be biological implications. Although the human memory system is by far a more complex mechanism than those that are outlined in this work, potential implications could be drawn from the present results. Different from machines human memory works often in a way that 
humans can remember unexpected events better. We usually can neither recall what we ate for lunch 78 days ago, nor precisely reconstruct the way we 
brushed our teeth last Monday. Events that are irrelevant or predictable from previous information or both are filtered away.

Also, human memory does not work from the first day, an effect known as childhood amnesia \cite{early}. A little child has early experiences that are not stored as repeatable events in later life. Rather it is plausible to assume that these early childhood experiences form a framework of expectations of usual daily events. 
Only the deviations from these expectations are stored and stay as memories at later stages of life. 

Another important hint that a mechanism similar to what is outlined in this paper is at work in the cortex are the ubiquitous predictive properties of the cortex 
recently have become common knowledge in the community
\cite{moshebar2011}.

\section*{Calculations for this manuscript are published online}
The numerical calculations have been performed by using the open source sagemath.org \cite{sagemath} package.
A part of the calculations are available under \cite{Fig6sgra}. 

\section{Acknowledgements}
The Ministry of Science and Technology (MOST) of Taiwan provided the budget for our project - project numbers : 103-2221-E-194 -039 and 102-2221-E-194 -050. 
Also thanks go to AIM-HI for various ways of support. I also thank Chris Shane for his cross reading.

\bibliographystyle{plain} 
\bibliography{ESN}

\end{document}